\documentclass[runningheads]{llncs}

\usepackage{eccv}

\usepackage{eccvabbrv}

\usepackage{fontspec}
\usepackage{graphicx}
\usepackage{booktabs}
\usepackage{float}
\usepackage{multirow}

\usepackage{pifont}       
\usepackage{caption}      
\usepackage{adjustbox}    
\usepackage[normalem]{ulem} 
\newcommand{\best}[1]{\textbf{#1}}
\newcommand{\second}[1]{\uline{#1}}
\newcommand{\dz}{\textcolor{gray}{\scriptsize(0.0)}}
\newcommand{\basecell}[1]{\shortstack[c]{#1\\\textcolor{gray}{\scriptsize(--)}}}
\newcommand{\dcell}[2]{\shortstack[c]{#1\\#2}}

\newcommand{\ddn}[1]{\textcolor{red!70!black}{\scriptsize(#1)}}   

\usepackage[pagebackref,breaklinks,colorlinks,citecolor=eccvblue]{hyperref}
\usepackage[table]{xcolor}

\usepackage{orcidlink}

\begin{document}



\author{Weijie Zhou\inst{1,2} \and
Xuantang Xiong\inst{3} \and
Zhenlin Hu\inst{5} \and
XiaoMeng Zhu\inst{4} \and
Chaoyang Zhao\inst{2} \and
Honghui Dong\inst{1} \and
Zhengyou Zhang\inst{3} \and
Ming Tang\inst{2} \and
Jinqiao Wnag\inst{2} }

\authorrunning{F.~Author et al.}

\institute{
Beijing Jiaotong University \and
Foundation Model Research Center, Institute of Automation, Chinese Academy of Sciences (CASIA) \and 
Tencent Robotics X \and
Department of Computer Science and Engineering, The Hong
Kong University of Science and Technology (HKUST) \and
Harbin Institute of Technology, Shenzhen \\
\email{\{chaoyang.zhao, jqwang\}@nlpr.ia.ac.cn} \\
\url{https://github.com/jetteezhou/Listening-with-the-Eyes} \\
}

\title{Listening with the Eyes: Benchmarking Egocentric Co-Speech Grounding across Space and Time} 
\titlerunning{Listening with the Eyes}

\maketitle
\begin{figure}[H]
    \centering
    \includegraphics[width=0.9\textwidth]{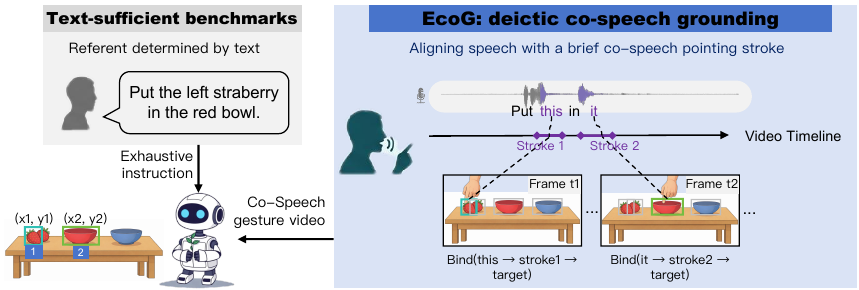} 
    \caption{\textbf{From text-sufficient grounding to deictic co-speech event binding.}
    \textbf{Left:} In many existing embodied/grounding benchmarks, the instruction is semantically exhaustive (e.g., attributes and spatial relations), so the correct referent can be inferred from text alone and co-speech gesture video is largely optional. 
    \textbf{Right:} \textbf{EcoG} models natural deictic collaboration, where utterances are intentionally underspecified (e.g., “put \emph{this} in \emph{it}”) and become solvable only by aligning each deictic phrase to a brief co-speech pointing \emph{stroke} on the video timeline. 
    Successful EcoG grounding requires \emph{within-clip event assignment}: binding each phrase to the correct stroke, then producing an executable intent for every step (\textit{What} target, \textit{Where} actionable 2D point, and \textit{When} stroke time).}
    \label{fig:paradigm_shift}
\end{figure}

\begin{abstract}
In situated collaboration, speakers often use intentionally underspecified deictic commands (e.g., ``pass me \textit{that}''), whose referent becomes identifiable only by aligning speech with a brief co-speech pointing \emph{stroke}.
However, many embodied benchmarks admit language-only shortcuts, allowing MLLMs to perform well without learning the \emph{audio--visual alignment} required by deictic interaction.
To bridge this gap, we introduce \textbf{Egocentric Co-Speech Grounding (EcoG)}, where grounding is executable only if an agent jointly predicts \textit{What}, \textit{Where}, and \textit{When}.
To operationalize this, we present \textbf{EcoG-Bench}, an evaluation-only bilingual (EN/ZH) diagnostic benchmark of \textbf{811} egocentric clips with dense spatial annotations and millisecond-level stroke supervision.
It is organized under a \textbf{Progressive Cognitive Evaluation} protocol.
Benchmarking state-of-the-art MLLMs reveals a severe executability gap: while human subjects achieve near-ceiling performance on EcoG-Bench (\textbf{96.9\%} strict Eco-Accuracy), the best native video-audio setting remains low (Gemini-3-Pro: \textbf{17.0\%}).
Moreover, in a diagnostic ablation, replacing the native video--audio interface with timestamped frame samples and externally verified ASR (with word-level timing) substantially improves the same model (\textbf{17.0\%}$\to$\textbf{42.9\%}).
Overall, EcoG-Bench provides a strict, executable testbed for event-level speech--gesture binding, and suggests that multimodal interfaces may bottleneck the observability of temporal alignment cues, independently of model reasoning.

  \keywords{Embodied AI \and Multimodal LLMs \and Spatiotemporal Grounding \and Co-speech Gestures}
\end{abstract}

\section{Introduction}
\label{sec:intro}

Human communication in situated collaboration follows the ``Principle of Least Effort'' \cite{human_behavior,situated_interaction2}:
rather than providing exhaustive descriptions (attributes, locations), speakers frequently use underspecified deictic utterances (e.g., ``give me \textit{that}'') and let co-speech gestures resolve reference \cite{cospeech,co-speech_gesture2,co-speech_gesprompt}.
In these interactions, the key information is carried by a short \emph{event}---the temporal coupling between a deictic word/phrase and the peak of a pointing gesture (gesture \emph{stroke})---which establishes joint attention \cite{co-speech_attention,co-speech_attention2,co-speech_attention3}. 
Crucially, without this timing binding, the same deictic words can match multiple plausible candidates in the scene.
To act as collaborative partners, embodied agents must therefore perform \emph{event-level} speech--gesture binding.

Despite the central role of co-speech gestures in collaboration, existing embodied and grounding benchmarks largely remain \emph{text-sufficient} (e.g., ``pick up the red apple on the left''), where language alone nearly determines the target \cite{refcoco,alfred,vln_bench1,vln_bench2,vln_bench3,vla_bench1,vla_bench2,vla_bench3}.
They also seldom require \emph{time-resolved} commitments: the decisive cue is a brief gesture \emph{stroke}, yet stroke-level temporal supervision and evaluation are typically absent. In multi-referent commands, this becomes a multi-event intent chaining problem.
The agent must map each deictic phrase to the correct stroke among several closely spaced events; a single mis-assignment can cascade.
This leaves an open question: can current MLLMs reliably perform such look-while-listen alignment under native video--audio interfaces?
We use EcoG-Bench to benchmark event-level speech--gesture binding under native video--audio inputs with strict executability-oriented metrics, and to diagnose whether multimodal input pipelines expose usable temporal anchors.

To evaluate this capability, we introduce \textbf{Egocentric Co-Speech Grounding (EcoG)}, which uses deictic language and requires resolving reference from spatiotemporal cues in egocentric video.
Given a clip with speech, the agent must produce an executable intent for each referent as a triplet:
\textit{What} (semantic referent), \textit{Where} (a precise 2D target point), and \textit{When} (a millisecond timestamp within the disambiguating gesture-stroke window).
The core challenge is \emph{event-level} fine-gained speech--gesture binding.

We build \textbf{EcoG-Bench}, a bilingual (EN/ZH) evaluation-only benchmark of \textbf{811} egocentric clips with dense spatial labels and millisecond stroke windows.
It follows a \textbf{Progressive Cognitive Evaluation} protocol that scales from single-event binding to within-clip event assignment and multi-step intent chaining.
This hierarchy makes error accumulation explicit as EcoG scales from single-event binding to \textit{within-clip event assignment} (multiple deictic cues) and \textit{multi-event intent chaining} under strict spatiotemporal constraints.

We evaluate EcoG with strict \textit{What/Where/When} metrics, including conjunctive \textbf{Eco-Accuracy} ($Acc_{eco}$) that requires all dimensions to be correct.
EcoG-Bench is well-posed for humans (near-ceiling \textbf{96.9\%} $Acc_{eco}$), yet remains challenging for modern MLLMs: under native video-audio interfaces, strict executability is low (e.g., \textbf{Gemini-3-Pro: 17.0\%} $Acc_{eco}$), and sequence success collapses as referents compose over time.
EcoG-Bench also supports \emph{system-level} diagnosis beyond model ranking.
For the same Gemini model, a scaffolded \textbf{multi-image + ASR} probe---providing sampled frames with timestamps and externally verified ASR with word-level timing---substantially improves strict grounding (\textbf{17.0\%$\to$42.9\%} $Acc_{eco}$).
This diagnostic probe is not information-equivalent to native inputs and is excluded from leaderboard comparison, but the large gain suggests that native interfaces may not reliably surface alignment cues.
More broadly, EcoG-Bench turns a core ingredient of human collaboration---binding deictic language to transient visual events---into a strict and executable evaluation target.
We hope it will facilitate progress on both model-level event binding and interface-level temporal alignment in next-generation embodied systems.

Our contributions are three-fold:
\begin{itemize}
  \item \textbf{Task.} We introduce \textbf{EcoG}, requiring executable \textit{What/Where/When} predictions for deictic co-speech commands.
  \item \textbf{Benchmark.} We build \textbf{EcoG-Bench} (811 clips, EN/ZH) with instance-level spatial targets and millisecond stroke windows under a progressive L1--L4 protocol.
  \item \textbf{Findings \& diagnosis.} We reveal a large executability gap for state-of-the-art MLLMs under native video--audio inputs, and show that adding explicit temporal anchors in the input interface can substantially improve event binding in a diagnostic setting.
\end{itemize}

\section{Related Work}

\subsection{Multimodal Instruction Following for Embodied Agents}

Embodied AI has advanced from language-guided navigation to long-horizon manipulation and interaction~\cite{Anderson2018R2R,Ku2020RxR,alfred,Padmakumar2022TEACh}, and recent systems couple large language models with multimodal perception for decision making~\cite{Driess2023PaLME,Brohan2023RT2,openvla}. 
Most benchmarks assume semantically specific instructions, under-testing deictic collaboration where reference must be resolved from co-speech events~\cite{alfred, physvlm-avr, physvlm}. EcoG targets executable intent grounding from stroke-level speech--gesture binding.

\subsection{Visual Grounding and Referring Expression Comprehension}

Referring expression comprehension and visual grounding localize objects specified by language, typically with attribute-rich descriptions in images (e.g., RefCOCO/+/g~\cite{Kazemzadeh2014ReferItGame}, Flickr30k Entities~\cite{flickr30k}). 
Video extensions further study temporal grounding by aligning sentences to segments~\cite{Gao2017TALL,activitynet,Tang2021HCSTVG}. 
EcoG differs from prior temporal grounding by using deictic-dominant language and requiring an \emph{actionable} commitment: a precise 2D target on the final frame \emph{and} a millisecond timestamp within the disambiguating stroke window.

\subsection{Egocentric Perception and Co-Speech Gestures}

Egocentric datasets such as EPIC-KITCHENS~\cite{epic-kitchens}, Ego4D~\cite{Grauman2022Ego4D}, and Ego-Exo4D~\cite{Grauman2024EgoExo4D} advance first-person action and hand--object modeling, but mainly study \emph{wearer-centric} activity rather than \emph{partner-centric} intent.
HRI work on communicative gestures and joint attention~\cite{cospeech,co-speech_attention,co-speech_attention2,co-speech_attention3} and YouRefIt~\cite{Chen2021YouRefIt} move toward referencing, yet deictic-heavy instructions with millisecond-level stroke supervision remain scarce.
Table~\ref{tab:dataset_comparison} summarizes these differences: EcoG pairs deictic speech with millisecond stroke windows and instance-level spatial targets for executable grounding.

\begin{table}[t]
\centering
\caption{Comparison of EcoG with related grounding and interaction datasets. EcoG uniquely integrates egocentric vision, audio, deictic ambiguity, and precise gesture stroke annotations.}
\resizebox{\textwidth}{!}{%
\begin{tabular}{l|c|c|c|c|c}
\toprule
\textbf{Dataset} & \textbf{Egocentric} & \textbf{Audio} & \textbf{Deictic Dominance} & \textbf{Gesture Stroke} & \textbf{Task Annotation} \\ 
\midrule
RefCOCO/+/g~\cite{Kazemzadeh2014ReferItGame} & \ding{55} & \ding{55} & \ding{55} & \ding{55} & Box/Mask \\
HC-STVG~\cite{Tang2021HCSTVG} & \ding{55} & \ding{51} & \ding{55} & \ding{55} & Time + Box \\
YouRefIt~\cite{Chen2021YouRefIt} & \ding{51} & \ding{51} & \ding{55} & \ding{51} (Gesture, coarse) & Box/Mask \\
Ego4D (FHO)~\cite{Grauman2022Ego4D} & \ding{51} & \ding{51} & \ding{55} & \ding{55} & Time + Box \\
ToG-Bench~\cite{tog_bench} & \ding{51} & \ding{55} & \ding{55} & \ding{55} & Time + Box \\
\rowcolor{blue!10} \textbf{EcoG (Ours)} & \ding{51} & \ding{51} & \ding{51} & \ding{51} (Stroke, ms-level) & Class + Mask + Time \\
\bottomrule
\end{tabular}%
}
\label{tab:dataset_comparison}
\end{table}

\subsection{Cognitive Evaluation of Multimodal LLMs}

General MLLM benchmarks (e.g., MMBench~\cite{Liu2023MMBench}, MMMU~\cite{Yue2024MMMU}, POPE~\cite{Li2023POPE}, MathVista~\cite{mathvista}) mainly adopt VQA-style formats to probe perception and reasoning, but rarely enforce \emph{executable} spatiotemporal commitments.
EcoG-Bench instead evaluates strict conjunctive correctness over \textit{What/Where/When} and progressively composes multiple referents (L1--L4), directly targeting the event-binding failure modes that limit embodied collaboration.

\section{The EcoG-Bench: Task, Data, and Metrics}
\label{sec:benchmark}

This section defines the \textbf{Egocentric Co-Speech Grounding (EcoG)} task and presents \textbf{EcoG-Bench}, a diagnostic benchmark designed to stress-test \emph{online event-level} speech--gesture binding in situated collaboration.
EcoG-Bench targets executable co-speech grounding: success requires correct semantics, actionable localization, and alignment to the disambiguating stroke event.
We describe (i) the task formulation and output structure, (ii) the bilingual data construction and annotation pipeline, (iii) the Progressive Cognitive Evaluation protocol (L1--L4), and (iv) strict metrics over \textit{What/Where/When}.

\subsection{Task Formulation}
\label{subsec:task_formulation}

\begin{figure}[t]
    \centering
    \includegraphics[width=0.95\textwidth]{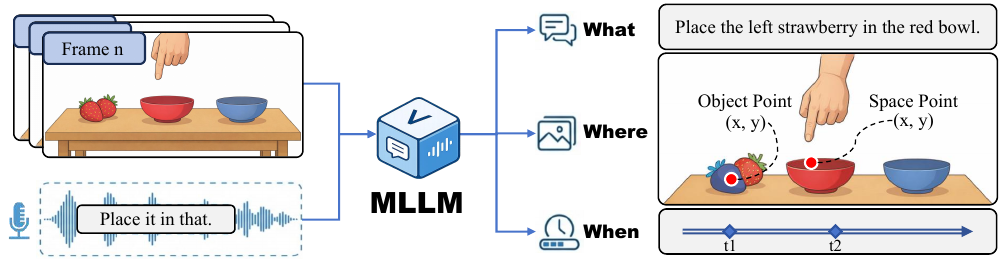}
    \caption{\textbf{EcoG task overview.}
    Given an egocentric video clip with synchronized audio, the model must ground each deictic referent in the instruction by outputting an ordered list of triplets: \textbf{What} (an index in a clip-specific closed-set of candidate options), \textbf{Where} (a 2D point on the last frame, ensuring an actionable “landing point”), and \textbf{When} (an integer timestamp in milliseconds from clip start that must fall inside the annotated gesture-stroke window that disambiguates the referent).}
    \label{fig:task}
\end{figure}

EcoG models natural situated collaboration where language is underspecified and reference must be resolved through co-speech gestures.
We denote the input as an egocentric clip $\mathcal{V}=\{v_t\}_{t=1}^{T}$ with its synchronized audio $\mathcal{A}$. The input specifies the number of referents $K$ and their execution order. The spoken instruction contains $K$ deictic referents $\mathcal{P}=\{p_1,\ldots,p_K\}$.

EcoG considers both single-step ($K{=}1$) and compositional multi-step instructions ($K\in\{2,3,4\}$). The goal is to ground each deictic referent $p_k$ into an \emph{executable} spatiotemporal intent.
We denote the prediction for $\mathcal{P}=\{p_1,\ldots,p_K\}$ as $\mathcal{Y}$, an ordered list of $K$ grounding triplets:
\begin{equation}
\mathcal{Y}=\Phi(\mathcal{V},\mathcal{A})
=\big[(c_1,s_1,\tau_1),\ldots,(c_K,s_K,\tau_K)\big],
\end{equation}
where $c_k \in \{1,\ldots,M\}$ is the predicted option index in a clip-specific closed-set candidate list (\textit{What}, $M{=}6$--$8$),
$s_k$ is a 2D point \textbf{on the last frame of the clip} (\textit{Where}). 
During curation, we ensure the intended target is visible on the last frame, so $s_k$ serves as an actionable landing point.
Each referent is typed as either a \texttt{target\_object} or a \texttt{spatial\_affordance}, which affects the spatial evaluation criterion. For the $k$-th referent, the triplet components are defined as:
\begin{itemize}
    \item \textbf{What ($c_k$):} The semantic category or description of the target object (\eg, ``Screwdriver'').
    \item \textbf{Where ($s_k$):} The precise spatial localization coordinate $s_k=(x,y)$ within the visual frame.
    \item \textbf{When ($\tau_k$):} The \textbf{predicted timestamp} of the critical temporal cue. Specifically, $\tau_k$ is an integer millisecond timestamp that should fall within the annotated gesture-stroke window that disambiguates $p_k$.
\end{itemize}

This serialized formulation scales naturally from atomic commands ($K{=}1$) to multi-step intents ($K\in\{3,4\}$).
The core difficulty is \textbf{event-level cross-modal binding}: the model must associate a deictic speech cue (with precise timing) to a brief gesture stroke and infer the intended target under egocentric clutter and viewpoint.
Importantly, EcoG also enables diagnosing whether different \emph{multimodal input pipelines} preserve such binding cues reliably (e.g., native video-audio vs. structured frames+ASR).

\subsection{Data Construction and Statistics}

EcoG-Bench is curated as a \textbf{diagnostic} benchmark for fine-grained co-speech grounding.
We follow three design principles:
\textbf{Situated Interaction} (captured in real collaborative workflows),
\textbf{Deictic Dominance} (instructions are intentionally underspecified and gesture-dependent),
and \textbf{Full-Stack Supervision} (aligned semantic, spatial, and millisecond-level temporal annotations).
EcoG-Bench is bilingual (EN/ZH) and evaluation-only to reduce contamination.

\begin{figure}[t]
    \centering
    \includegraphics[width=1\textwidth]{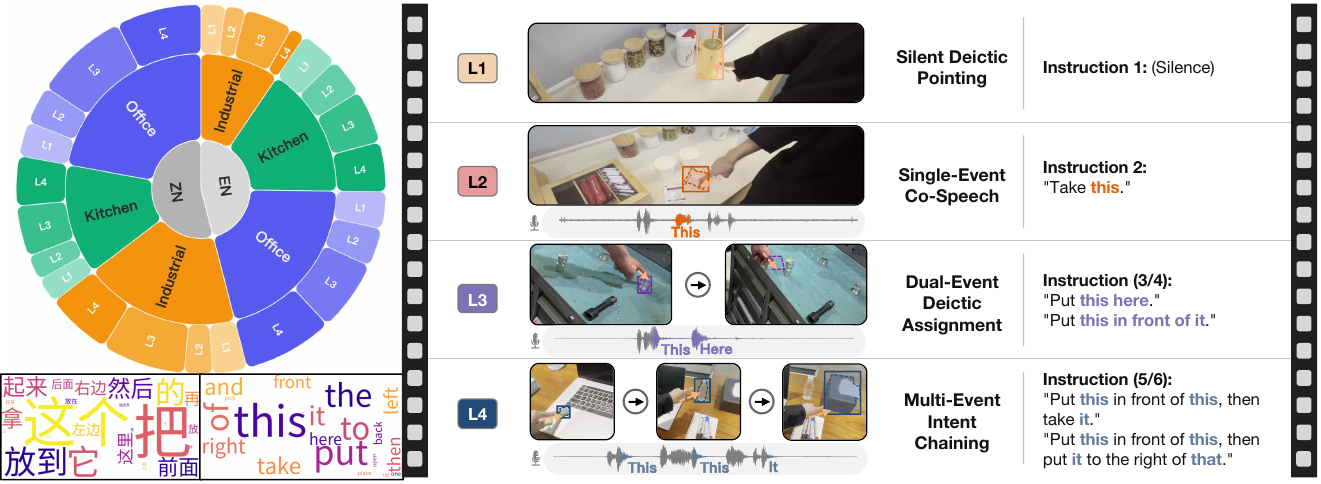} 
    \caption{\textbf{Progressive Cognitive Evaluation protocol and dataset composition.} EcoG-Bench organizes 811 egocentric clips (EN/ZH) into four levels with increasing compositionality and event-assignment difficulty: \textbf{L1} silent deictic pointing (K=1), \textbf{L2} single-event co-speech binding (K=1), \textbf{L3} dual-event deictic assignment (K=2), and \textbf{L4} multi-event intent chaining (K=3–4). The figure illustrates the corresponding instruction templates and the increasing requirement to assign each deictic phrase to the correct within-clip gesture stroke.}
    \label{fig:dataset}
\end{figure}

\textbf{Collection Protocol: Dyadic Situated Collaboration.}
We record human--human collaborative interactions where one participant issues directives (``User'') and the other follows (``Agent'').
To enforce \textbf{deictic dominance}, we adopt a strict \emph{No Explicit Description} rule: users are instructed to avoid exhaustive attributes/locations and instead use deictic phrases (e.g., ``this/that/here/there'') accompanied by pointing gestures.
As shown in Figure~\ref{fig:dataset}, EcoG-Bench contains \textbf{811} curated clips (\textbf{367} EN, \textbf{444} ZH) spanning three domains: \textit{Industrial}, \textit{Kitchen}, and \textit{Office}, and covers 6 instruction templates (Instruction1--6) grouped into four cognitive levels (L1--L4).
For \textit{What} evaluation, each clip is paired with a clip-specific closed-world candidate set of \textbf{scene-visible} options ($M{=}6$--$8$) for unambiguous scoring. 
Options are \textbf{text-only} object descriptions; visually similar instances (e.g., two identical cups) are distinguished as different options by their \textbf{unique spatial instances/locations} in the scene. 
The option order is randomized per clip to reduce ordering biases.

\textbf{Full-Stack Annotation Pipeline (Semantic--Spatial--Temporal).}
We build reproducible supervision for \textit{What/Where/When} via a three-stage pipeline:
\begin{enumerate}
    \item \textbf{Semantic labeling (\textit{What}).}
    Each referent is mapped to a clip-specific closed-world option set ($M{=}6$--$8$ candidates) and labeled with the correct option.
    This avoids ambiguity from open-vocabulary synonyms during evaluation.

    \item \textbf{Spatial grounding (\textit{Where}).}
    Annotators click a pixel on the \textbf{last frame} to indicate the target.
    For object referents, we generate an instance mask using \textbf{SAM-3} \cite{sam3} seeded by the click point, followed by manual verification for small/occluded objects.
    For non-object referents (e.g., placement regions), we annotate a point target without a mask.

    \item \textbf{Temporal grounding (\textit{When}).}
    We first transcribe speech using \textbf{Fun-ASR} and manually verify the transcript.
    Word-level timestamps (ms) provide precise time anchors.
    We then align each referent to its deictic word/phrase time span (LLM-assisted and human-verified) and annotate a \textbf{gesture stroke window} by directly labeling the temporal interval $[t^{k}_{start}, t^{k}_{end}]$ that brackets the visually observed gesture peak used for disambiguation. 
\end{enumerate}
Finally, all annotations go through a multi-reviewer QA process.
We require inter-annotator agreement above standard thresholds (Cohen's $\kappa > 0.80$) and obtain $\kappa=0.84$ for spatial labels and $\kappa=0.87$ for classification labels in our released set (see Supp.; annotation guidelines, interface, QA workflow, and agreement analysis are provided there).

\textbf{Dataset Statistics.}
As shown in Figure~\ref{fig:dataset}, EcoG-Bench contains \textbf{811} egocentric clips (4--12s), including \textbf{367} English and \textbf{444} Chinese instances, spanning \textit{Industrial}, \textit{Kitchen}, and \textit{Office} domains and covering L1--L4 with 6 instruction templates (Instruction1--6).

\subsection{Progressive Cognitive Evaluation Protocol}
\label{subsec:hierarchy}

To diagnose failure modes beyond single-step perception, we propose a \textbf{Progressive Cognitive Evaluation} protocol that increases compositionality along two axes: the number of referents $K$ and the need for \textit{within-clip event assignment} (mapping each deictic cue to the correct gesture stroke) and \textit{multi-event intent chaining} under strict executability.
EcoG-Bench instantiates this protocol using 6 instruction templates (Instruction1--6), grouped into four levels (L1--L4):

\textbf{Level 1 (L1): Silent Deictic Pointing (Instruction1, $K{=}1$).}
\textit{Focus: Pure visual deixis—pointing geometry \emph{and} temporal stroke localization under egocentric viewpoint (no speech).}
The user points to a target without speech (Instruction1: \textit{silence}).
The agent must infer \textit{What}, \textit{Where}, and \textit{When}, where \textit{When} corresponds to the visually observed pointing stroke window.

\textbf{Level 2 (L2): Single-Event Co-Speech Binding (Instruction2, $K{=}1$).}
\textit{Focus: Event-level audio--visual binding between a deictic word/phrase and a single gesture stroke.}
The user issues a single deictic command with one referent (Instruction2: ``Take \underline{this}.'') accompanied by a pointing gesture.
The agent must predict \textit{What/Where/When}, where \textit{When} is scored by whether the predicted timestamp falls inside the annotated gesture stroke window for the deictic phrase.
This level isolates the core alignment capability and is sensitive to whether the input pipeline preserves reliable timing anchors.

\textbf{Level 3 (L3): Dual-Event Deictic Assignment (Instruction3--4, $K{=}2$).}
\textit{Focus: Within-clip event assignment across two deictic cues (word/phrase $\leftrightarrow$ stroke), plus spatial constraints (placement/relation).}
L3 covers two templates:
(i) \textbf{Instruction3} (Instruction3: ``Put \underline{this} \underline{here}.'', object placement: \texttt{target\_object} $\rightarrow$ \texttt{spatial\_affordance}),   and (ii) \textbf{Instruction4} (Instruction4: ``Put \underline{this} in front of \underline{this}.'', relational placement between two grounded referents).
In both cases, the key difficulty is assigning each deictic cue to the correct gesture stroke within the same clip.

\textbf{Level 4 (L4): Multi-Event Intent Chaining (Instruction5--6, $K\in\{3,4\}$).}
\textit{Focus: Ordered event chaining with referential state tracking under strict executability.}
L4 contains two multi-step templates: \textbf{Instruction5} (3 referents; Instruction5: ``Put \underline{this} to the right of \underline{this}, and take \underline{it}.'') and   \textbf{Instruction6} (4 referents; Instruction6: ``Put \underline{this} to the right of \underline{this}, then put \underline{this} to the left of \underline{that}.''). These templates require multi-event intent chaining and explicit referential state tracking across successive deictic cues.

\subsection{Evaluation Metrics}
\label{subsec:metrics}

EcoG-Bench evaluates \textit{What/Where/When} with component metrics and strict composite/sequence metrics.
We emphasize \textbf{executability}: a prediction only counts as correct when it is semantically correct, spatially actionable, and temporally aligned to the disambiguating gesture event.

\textbf{1) Component Metrics.}
For each referent, we compute:
\begin{itemize}
    \item \textbf{Classification Accuracy ($Acc_{cls}$, \textit{What}).}
    A prediction is correct if $c_{\text{pred}} = c_{\text{gt}}$ (the correct index in the clip-specific candidate set).

    \item \textbf{Spatial Accuracy ($Acc_{s}$, \textit{Where}).}
    Predictions are evaluated on the \textbf{last frame}.
    If an instance mask $\mathcal{M}_{gt}$ is available (object referents), $Acc_s{=}1$ iff $s_{\text{pred}}\in\mathcal{M}_{gt}$.
    If the referent is a \texttt{spatial\_affordance} without a mask, we score by a pixel-distance threshold:
    $Acc_s{=}1$ iff $\lVert s_{\text{pred}} - s_{\text{gt}}\rVert_2 < \delta$,
    with $\delta{=}100$ px (ranking-stable for $\delta\in[100,150]$; see Supp.).

    \item \textbf{Temporal Accuracy ($Acc_{t}$, \textit{When}).}
    For L1--L4, each referent has an annotated gesture stroke window $[t^{k}_{start},t^{k}_{end}]$ in milliseconds (derived from ASR-aligned deictic phrase time spans and verified with the visual gesture).
    A prediction is correct if $\tau_{\text{pred}}\in[t^{k}_{start},t^{k}_{end}]$.
\end{itemize}

\textbf{2) Composite Metric: Eco-Accuracy ($Acc_{eco}$).}
EcoG requires predictions to be \emph{jointly} correct in semantics, actionable localization, and event timing.
We therefore define, for all levels (L1--L4):
\begin{equation}
Acc_{eco}(\text{referent}) \;=\; \mathbb{I}\!\left(Acc_{cls}=1 \;\land\; Acc_{s}=1 \;\land\; Acc_{t}=1\right).
\end{equation}
This strict conjunction aligns with executability: a model must identify the correct referent (\textit{What}), point to the correct actionable location (\textit{Where}), and bind to the correct disambiguating event (\textit{When}).

\textbf{3) Sequence Metric: $Acc_{seq}$.}
To reflect real executability under multi-intent instructions (L3: $K{=}2$; L4: $K\in\{3,4\}$ for Instruction5--6),
we define sequence-level success with an all-or-nothing criterion:
a clip is correct iff \emph{every} referent in the instruction attains $Acc_{eco}{=}1$.
For a dataset of $N$ clips:
\begin{equation}
Acc_{seq} = \frac{1}{N}\sum_{i=1}^{N}\prod_{k=1}^{K_i}\mathbb{I}\Big(\text{referent}_{i,k}\ \text{is Eco-correct}\Big).
\label{eq:seq_metric}
\end{equation}
This instance-level logical AND captures error cascading in compositional grounding and makes EcoG-Bench sensitive to small mis-calibrations in spatiotemporal binding, which are often hidden by marginal metrics.

\section{Experiments and Analysis}
\label{sec:experiments}

We benchmark state-of-the-art MLLMs on EcoG-Bench under strict executability-oriented metrics.
Unlike text-sufficient grounding, EcoG becomes non-executable as soon as the model loses the audio--visual alignment that identifies the correct stroke event, even if object recognition is strong.
We evaluate state-of-the-art MLLMs on EcoG-Bench under strict executability metrics, and run a controlled input-stack diagnostic that varies only the multimodal interface.

\subsection{Experimental Setup}
\label{subsec:exp_setup}

\textbf{Models.}
We evaluate representative MLLMs spanning:
(i) \textbf{Native Omni} models that ingest raw video files with audio end-to-end, and
(ii) \textbf{Vision-Language (VL)} models that operate on sampled frames plus text.
For all models, the prompt provides the number of referents $K$ and requires an ordered list of $K$ triplets in JSON.

\textbf{Diagnostic probe on the multimodal interface.}
To test whether EcoG failures stem from the input pipeline (temporal cue exposure) rather than model weights alone,
we run a \textbf{diagnostic ablation} on several omni models (with Gemini-3-Pro/Flash as the main focus) under two interfaces:
\begin{itemize}
    \item \textbf{Video-Omni (native):} the default end-to-end interface that takes a single video file with audio.
    \item \textbf{Images + ASR (scaffolded, diagnostic):} uniformly sampled frames augmented with per-frame timestamps, plus externally produced ASR transcripts (Fun-ASR) that are manually verified and include word-level begin/end timestamps.
\end{itemize}
We keep the option sets, prompts, output schema, and metrics identical, and vary only the input representation.

\textbf{Input standardization.}
Spatial coordinates are evaluated on the last frame (targets are curated to be visible; see Sec.~\ref{subsec:task_formulation}).

\textbf{Video-Omni pipeline.}
For native omni models, we provide the raw video file with its audio track and the option list.
No external ASR transcript or explicit frame timestamps are injected.

\textbf{Frame-based (VL) pipeline.}
For frame-based evaluation, we uniformly sample frames at \textbf{2 fps} and attach a \texttt{<timestamp\_ms>} tag to each frame.
We provide a manually verified ASR transcript as plain text (without word-level timing) to avoid injecting explicit temporal anchors in the standard setting.
We include a frame-sampling sensitivity study in the Supplementary Material.

\textbf{Images+ASR (diagnostic) pipeline.}
In the diagnostic ablation (Sec.~\ref{sec:input_stack}), we additionally augment the same frame stream with ASR word-level begin/end timestamps via \texttt{<asr\_data>} (Fun-ASR, manually verified; see Supp.).

\textbf{Coordinate convention.}
Some models return points in different orders (e.g., Gemini uses \texttt{[y,x]}).
We normalize all predictions to a unified \texttt{[x,y]} convention before scoring.

\textbf{Decoding and validity.}
We use deterministic decoding (temperature $=0$).
Outputs must be valid JSON; unparsable responses are scored as incorrect for all metrics,
consistent with executability.

\begin{table*}[t]
\centering
\caption{\textbf{EcoG-Bench results under each model class's native interface.}
We report referent-level Eco-Accuracy ($Acc_{eco}$), clip-level sequence success ($Acc_{seq}$), and classification accuracy ($Acc_{cls}$) across levels L1--L4 and overall.
$Acc_{eco}$ is conjunctive: for all levels, a referent is correct only if \textit{What}$\land$\textit{Where}$\land$\textit{When} are correct.
Bold/underline indicate the best/second result \emph{per column} within each model family (Omni vs.\ VL), excluding Human.}
\setlength{\tabcolsep}{4.2pt}
\renewcommand{\arraystretch}{1.15}

\resizebox{\textwidth}{!}{%
\begin{tabular}{l|ccc|ccc|ccc|ccc|ccc}
\toprule
\multirow{2}{*}{\textbf{Model}} &
\multicolumn{3}{c|}{\textbf{L1}} &
\multicolumn{3}{c|}{\textbf{L2}} &
\multicolumn{3}{c|}{\textbf{L3}} &
\multicolumn{3}{c|}{\textbf{L4}} &
\multicolumn{3}{c}{\textbf{Overall}} \\
& eco & seq & cls & eco & seq & cls & eco & seq & cls & eco & seq & cls & eco & seq & cls \\
\midrule

\rowcolor{blue!10}
\textbf{Human Performance}
& 98.2 & 98.2 & 99.1
& 97.5 & 97.5 & 98.8
& 97.2 & 96.7 & 98.5
& 96.4 & 95.2 & 98.1
& 96.9 & 96.2 & 97.3 \\
\midrule

\multicolumn{16}{l}{\textit{\textbf{Native Omni Models} (Video-Omni: video+audio)}}\\

Gemini-3-Pro \cite{gemini}
& \best{30.2} & \best{30.2} & \best{79.9}
& \best{29.2} & \best{29.2} & \best{71.5}
& \best{10.6} & \best{1.8}  & \second{51.9}
& \best{10.2} & \best{0.4}  & \second{64.5}
& \best{17.0} & \best{10.9} & \second{63.9} \\

Gemini-3-Flash \cite{gemini}
& \second{12.2} & \second{12.2} & \second{71.2}
& \second{10.2} & \second{10.2} & \second{69.3}
& \second{3.2}  & \second{0.7}  & \best{65.3}
& \second{6.6}  & 0.0           & \best{79.5}
& \second{7.0}  & \second{4.1}  & \best{71.4} \\

Qwen3-Omni-30B-A3B \cite{qwen3-omni}
& 3.6 & 3.6 & 40.3
& 0.7 & 0.7 & 48.9
& 0.0 & 0.0 & 23.9
& 0.0 & 0.0 & 19.1
& 0.7 & 0.7 & 29.5 \\

Qwen3-Omni-Flash \cite{qwen3-omni}
& 2.9 & 2.9 & 59.7
& 0.7 & 0.7 & 66.4
& 0.2 & 0.0 & 40.5
& 0.0 & 0.0 & 37.2
& 0.7 & 0.6 & 47.1 \\

MiniCPM-o-4.5 \cite{MiniCPM-o}
& 0.0 & 0.0 & 40.3
& 0.0 & 0.0 & 43.1
& 0.0 & 0.0 & 29.2
& 0.0 & 0.0 & 23.4
& 0.0 & 0.0 & 31.7 \\

Ming-Lite-Omni-1.5 \cite{ming}
& 0.0 & 0.0 & 30.2
& 0.0 & 0.0 & 31.4
& 0.0 & 0.0 & 13.9
& 0.0 & 0.0 & 5.0
& 0.0 & 0.0 & 16.9 \\

\midrule
\multicolumn{16}{l}{\textit{\textbf{Vision-Language Models} (2fps Frames + ASR)}}\\

Qwen3-VL-30B \cite{qwen3}
& \second{18.0} & \second{18.0} & 64.0
& \best{19.7} & \best{19.7} & \best{60.6}
& \best{8.5}  & \best{0.7}  & \second{34.3}
& \best{6.6}  & 0.0         & 25.6
& \best{11.4} & \best{6.7}  & 41.2 \\

Qwen3-VL-8B \cite{qwen3}
& \best{21.6} & \best{21.6} & \second{66.9}
& \second{16.1} & \second{16.1} & \second{59.9}
& \second{4.4}  & \second{0.4}  & 32.7
& 2.7           & 0.0           & \second{30.5}
& \second{8.8}  & \second{6.5}  & \second{42.5} \\

GPT-5-mini \cite{gpt5}
& 5.0 & 5.0 & \best{74.8}
& 2.9 & 2.9 & 56.9
& 2.8 & \second{0.4} & \best{38.4}
& \second{3.2} & 0.0 & \best{47.3}
& 3.3 & 1.5 & \best{50.5} \\

LLaVA-OV-1.5-8B \cite{llava-ov1.5}
& 0.0 & 0.0 & 17.3
& 1.5 & 1.5 & 14.6
& 0.2 & 0.0 & 23.1
& 0.1 & 0.0 & 15.2
& 0.3 & 0.2 & 18.2 \\

LLaVA-NeXT-Video-7B \cite{llavanext}
& 0.0 & 0.0 & 18.7
& 0.0 & 0.0 & 11.7
& 0.0 & 0.0 & 13.4
& 0.0 & 0.0 & 4.6
& 0.0 & 0.0 & 11.3 \\

\bottomrule
\end{tabular}}
\label{tab:main_results}
\end{table*}

\begin{figure}[t]
    \centering
    \includegraphics[width=1\textwidth]{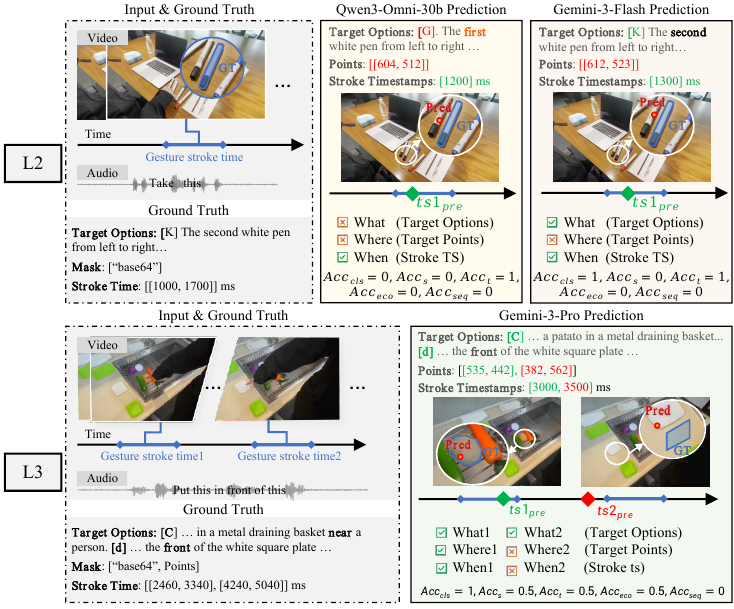} 
    \caption{\textbf{Qualitative results on EcoG-Bench.}
    Examples of model predictions versus ground truth under strict What/Where/When evaluation. The shown cases highlight typical failure modes: correct recognition but inaccurate pointing on small/occluded objects, and mis-binding a deictic phrase to a nearby (but incorrect) stroke event—both of which render the output non-executable under conjunctive Eco-Accuracy.}
    \label{fig:qualitative}
\end{figure}

\subsection{Main Results: The Gap in EcoG Task}
\label{subsec:main_results}

Table~\ref{tab:main_results} reports EcoG-Bench results under strict executability metrics (human evaluation protocol is detailed in the Supplementary Material).
We highlight three observations:

\textbf{(1) A large human--model gap under strict executability.}
Humans achieve near-ceiling performance across all levels ($96.9\%$ $Acc_{eco}$ and $96.2\%$ $Acc_{seq}$ overall), while state-of-the-art models remain far behind.
This confirms EcoG-Bench is a non-trivial stress test even for modern omni systems.

\textbf{(2) The largest compositional drop occurs from L2 to L3.}
Single-event co-speech grounding (L2) is already challenging but partially solvable
(e.g., Gemini-3-Pro reaches $29.2\%$ $Acc_{eco}$).
However, when instructions require \textit{within-clip event assignment} across multiple referents (L3) and \textit{multi-event intent chaining} (L4), performance collapses: Gemini-3-Pro drops to $10.6\%$ (L3) and $10.2\%$ (L4) $Acc_{eco}$,
and sequence success becomes near-zero ($1.8\%$ in L3; $0.4\%$ in L4).
This reflects a qualitatively harder regime: the model must solve \emph{dual-/multi-event deictic assignment} (which deictic cue binds to which gesture stroke) and then execute \textit{What/Where/When} commitments in the correct order. Under the conjunctive metric, a single mis-assigned timestamp or slight mis-calibration cascades into near-zero $Acc_{seq}$.

\textbf{(3) Semantic recognition $\neq$ executable grounding.}
Models can score reasonably on $Acc_{cls}$ but fail to produce actionable grounding.
For example, Gemini-3-Pro reaches $63.9\%$ $Acc_{cls}$ overall but only $17.0\%$ $Acc_{eco}$, motivating our input-stack diagnosis (Sec.~\ref{sec:input_stack}).

We further visualize typical failure cases in Fig.~\ref{fig:qualitative}, including accurate \emph{What} prediction but spatial misses on small/occluded targets and mis-binding deictic phrases to nearby (but incorrect) stroke events.

\subsection{Bottleneck Analysis: Decoupling \textit{What}, \textit{Where}, and \textit{When}}
\label{subsec:bottleneck}

\begin{figure}[t]
\centering
\includegraphics[width=1.0\linewidth]{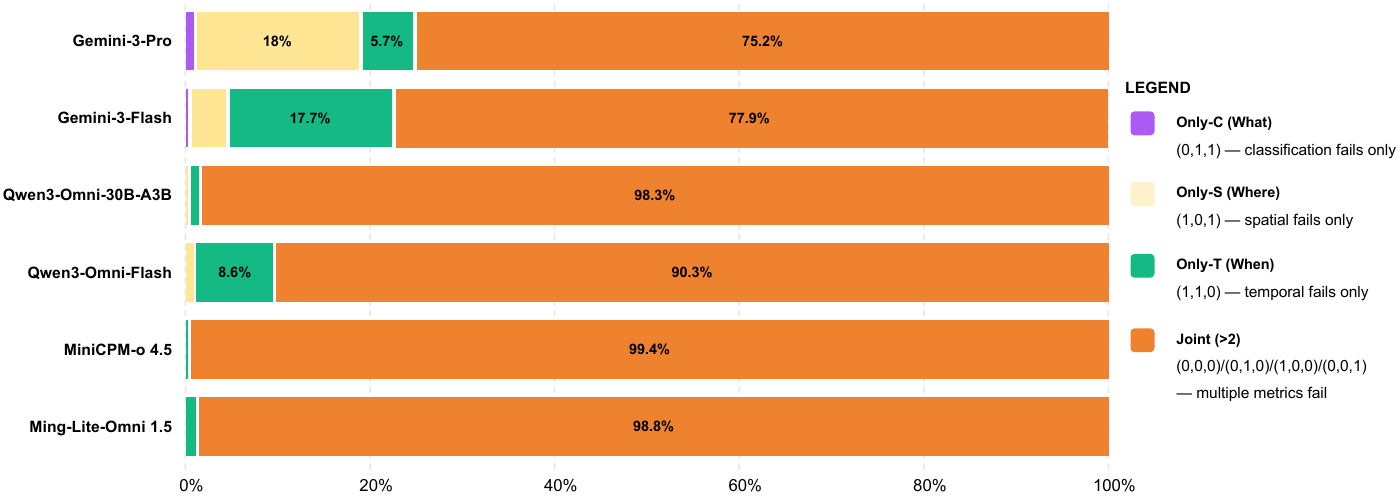}
\caption{\textbf{Failure bottleneck analysis of EcoG}. Breakdown of errors by which components of the executable grounding triplet fail (What, Where, When) and their combinations. Joint failures (e.g., Where+When) constitute a large portion of errors, indicating that EcoG difficulty is dominated by cross-modal event binding rather than isolated object classification alone.}
\label{fig:failure_bottleneck}
\end{figure}

To understand why strict $Acc_{eco}$ remains low, we decouple EcoG into
$Acc_{cls}$ (\textit{What}), $Acc_s$ (\textit{Where}), and $Acc_t$ (\textit{When})
and analyze failure bottlenecks.

\textbf{Unbalanced capability profiles.}
Across models, strong $Acc_{cls}$ does not imply strong executability:
a single spatial miss on a small object or a slightly mis-timed event prediction
invalidates the entire step under $Acc_{eco}$.

\textbf{Failures are predominantly joint rather than isolated.}
We further categorize errors by which sub-metrics fail. As shown in Figure~\ref{fig:failure_bottleneck}, a large fraction of samples fall into \emph{joint} failure categories (e.g., spatial+temporal failures, or all sub-metrics incorrect), indicating that EcoG’s difficulty is not a single missing skill, but the inability to robustly bind \textit{What/Where/When} into a single executable event.
Detailed bottleneck distributions and scene$\times$level heatmaps are provided in the Supplementary Material.

\subsection{Input-Stack Diagnosis: Images+ASR vs. Native Video-Omni}
\label{sec:input_stack}

\begin{table}[t]
\centering
\caption{Input-stack diagnosis (Overall): Images+ASR vs. native Video-Omni. We report strict executability metrics as \textit{Video-Omni}$\rightarrow$\textit{Images+ASR} (\(\Delta\)).}
\label{tab:input_stack_diagnosis}
\scriptsize
\setlength{\tabcolsep}{3.5pt}
\begin{tabular}{lcc}
\toprule
\textbf{Model} &
\textbf{\(Acc_{eco}\) (\%)} &
\textbf{\(Acc_{seq}\) (\%)} \\
\midrule
Gemini-3-Pro
& 17.0$\rightarrow$42.9 (+25.9)
& 10.9$\rightarrow$25.5 (+14.6) \\
Gemini-3-Flash
& 7.0$\rightarrow$48.1 (+41.1)
& 4.1$\rightarrow$30.8 (+26.7) \\
Qwen3-Omni-Flash
& 0.7$\rightarrow$9.3 (+8.6)
& 0.6$\rightarrow$6.5 (+5.9) \\
\bottomrule
\end{tabular}
\end{table}

Native omni models ingest raw video+audio end-to-end, but this is not the only way to supply multimodal information.
We therefore conduct a controlled \textbf{diagnostic-only} comparison on the \emph{same} omni model family,
varying \emph{only} the input pipeline: (i) native Video-Omni, and (ii) structured Images+ASR with per-frame time tags and word-level ASR timestamps.

\textbf{Gemini: structured Images + ASR dramatically improves strict grounding.}
As shown in Table~\ref{tab:input_stack_diagnosis}, for Gemini-3-Pro, $Acc_{eco}$ increases from $17.0\%$ (Video-Omni) to $42.9\%$ (Images+ASR),
and $Acc_{seq}$ increases from $10.9\%$ to $25.5\%$.
For Gemini-3-Flash, the improvement is even larger: $7.0\%\rightarrow48.1\%$ $Acc_{eco}$
and $4.1\%\rightarrow30.8\%$ $Acc_{seq}$.
Full per-level comparisons of the input-stack diagnosis are provided in the supplementary material. Notably, gains are consistent across L1--L4, suggesting that the structured pipeline provides a more reliable scaffold for both precise pointing and event timing.


\textbf{Takeaway.}
These results show that a scaffolded, time-anchored probe (word-level ASR timing + frame timestamps) can substantially improve strict EcoG executability for the same omni model. While this is not an information-equivalent comparison to native video--audio inputs, the large gain is consistent with a first-principles interpretation: explicitly anchored timestamps increase the \emph{observability} of word--stroke synchrony, which is a necessary cue for identifiable deictic grounding.

\subsection{Do Models Use Temporal Alignment Cues? Temporal Anchor Ablations}
\label{sec:temporal_anchor_ablation}

\begin{table*}[t]
\centering
\caption{\textbf{Temporal anchor ablations (diagnostic-only) under Images+ASR.}
We report \(Acc_t\) (temporal accuracy) and strict \(Acc_{eco}\) (executability) for each level and overall.
Each cell shows \textit{absolute score} (top) and \(\Delta\) vs. \textit{Full Anchors} (bottom), computed \textbf{within the same model}. Note that L1 contains no speech; thus ASR timing is unavailable and only frame timestamps can provide absolute temporal anchors.}
\label{tab:anchor_ablation}
\scriptsize
\setlength{\tabcolsep}{3.0pt}
\renewcommand{\arraystretch}{1.25}
\resizebox{\textwidth}{!}{
\begin{tabular}{ll|cc|cc|cc|cc|cc}
\toprule
\multirow{2}{*}{\textbf{Model}} & \multirow{2}{*}{\textbf{Condition}} &
\multicolumn{2}{c|}{\textbf{L1}} &
\multicolumn{2}{c|}{\textbf{L2}} &
\multicolumn{2}{c|}{\textbf{L3}} &
\multicolumn{2}{c|}{\textbf{L4}} &
\multicolumn{2}{c}{\textbf{Overall}} \\
& & \(t\) & \(eco\) & \(t\) & \(eco\) & \(t\) & \(eco\) & \(t\) & \(eco\) & \(t\) & \(eco\) \\
\midrule

& Full Anchors
& \basecell{94.2} & \basecell{66.9}
& \basecell{91.2} & \basecell{65.7}
& \basecell{77.1} & \basecell{40.0}
& \basecell{75.0} & \basecell{37.3}
& \basecell{81.8} & \basecell{48.1} \\
\rowcolor{gray!6}
\multirow{-3}{*}{\textbf{Gemini-3-Flash}}
& No Frame Timestamps
& \dcell{15.8}{\ddn{-78.4}} & \dcell{9.4}{\ddn{-57.5}}
& \dcell{85.4}{\ddn{-5.8}}  & \dcell{56.2}{\ddn{-9.5}}
& \dcell{70.1}{\ddn{-7.0}}  & \dcell{30.1}{\ddn{-9.9}}
& \dcell{67.8}{\ddn{-7.2}}  & \dcell{28.4}{\ddn{-8.9}}
& \dcell{62.7}{\ddn{-19.1}} & \dcell{30.4}{\ddn{-17.7}} \\
\rowcolor{gray!6}
& No Word-level ASR Timing
& \dcell{92.8}{\ddn{-1.4}} & \dcell{66.9}{\dz}
& \dcell{84.7}{\ddn{-6.5}} & \dcell{61.3}{\ddn{-4.4}}
& \dcell{75.2}{\ddn{-1.9}} & \dcell{38.2}{\ddn{-1.8}}
& \dcell{66.7}{\ddn{-8.3}} & \dcell{34.5}{\ddn{-2.8}}
& \dcell{77.2}{\ddn{-4.6}} & \dcell{45.9}{\ddn{-2.2}} \\

\midrule

& Full Anchors
& \basecell{88.5} & \basecell{59.0}
& \basecell{86.9} & \basecell{59.9}
& \basecell{72.9} & \basecell{34.5}
& \basecell{68.4} & \basecell{34.2}
& \basecell{76.5} & \basecell{42.9} \\
\rowcolor{gray!6}
\multirow{-3}{*}{\textbf{Gemini-3-Pro}} 
& No Frame Timestamps
& \dcell{20.9}{\ddn{-67.6}} & \dcell{14.4}{\ddn{-44.6}}
& \dcell{82.5}{\ddn{-4.4}}  & \dcell{43.1}{\ddn{-16.8}}
& \dcell{67.8}{\ddn{-5.1}}  & \dcell{18.5}{\ddn{-16.0}}
& \dcell{68.2}{\ddn{-0.2}}  & \dcell{17.7}{\ddn{-16.5}}
& \dcell{63.0}{\ddn{-13.5}} & \dcell{21.7}{\ddn{-21.2}} \\
\rowcolor{gray!6}
& No Word-level ASR Timing
& \dcell{88.5}{\dz}         & \dcell{56.8}{\ddn{-2.2}}
& \dcell{76.6}{\ddn{-10.3}} & \dcell{48.9}{\ddn{-11.0}}
& \dcell{62.9}{\ddn{-10.0}} & \dcell{26.2}{\ddn{-8.3}}
& \dcell{40.0}{\ddn{-28.4}} & \dcell{22.0}{\ddn{-12.2}}
& \dcell{62.5}{\ddn{-14.0}} & \dcell{34.0}{\ddn{-8.9}} \\

\bottomrule
\end{tabular}
}
\end{table*}
To isolate the effect of temporal anchoring (vs.\ added visual content), we run a controlled ablation with identical sampled frames and transcript text:
(i) \textbf{Full anchors} includes per-frame \texttt{<timestamp\_ms>} and word-level ASR begin/end times;
(ii) \textbf{No frame timestamps} removes \texttt{<timestamp\_ms>} while keeping frames and order unchanged;
(iii) \textbf{No word-level ASR timing} keeps transcript text but drops all word-level timing fields.

Tab.~\ref{tab:anchor_ablation} shows that removing per-frame timestamps causes the largest drop in temporal accuracy and strict executability, especially in L1.
This is expected because L1 is \emph{silent}: without speech, there are no ASR-derived anchors, and the model observes only an ordered frame sequence but must still output an absolute millisecond timestamp.
As a result, absolute-time prediction becomes underconstrained, leading to near-random \(Acc_t\) (and thus \(Acc_{eco}\)) in L1.

For L2--L4, speech provides an additional temporal reference: even without frame timestamps, word-level ASR timing can partially anchor when the deictic phrase occurs, so the degradation is smaller.
Conversely, removing word-level ASR timing consistently hurts L2--L4 (with no effect in L1), indicating that ASR timing mainly helps align deictic words to gesture strokes.
Overall, the ablation suggests that per-frame timestamps are critical for \emph{absolute-time calibration}, while word-level ASR timing further improves event-level speech--gesture binding.

\section{Conclusion}

We introduce EcoG and EcoG-Bench to evaluate executable, event-level co-speech grounding in egocentric collaboration. EcoG requires strict binding across \textbf{What/Where/When}: because deictic language is intentionally underspecified, correct grounding often depends on aligning speech to the correct gesture \textbf{stroke} in time. EcoG-Bench scales from single-event binding to within-clip multi-event assignment and intent chaining, evaluated with strict $\mathrm{Acc}_{eco}/\mathrm{Acc}_{seq}$.
Experiments reveal a large human--model gap and a sharp L2$\rightarrow$L3 compositional cliff: once multiple referents appear, the problem becomes within-clip event assignment, and small spatial or temporal errors quickly cascade to sequence failure. EcoG-Bench also enables system-level diagnosis beyond model weights. In a diagnostic-only ablation, adding explicit temporal anchors (multi-image input with word-timed ASR) substantially improves strict executability (e.g., $17.0\%\rightarrow 42.9\%~\mathrm{Acc}_{eco}$ for Gemini-3-Pro), suggesting current native video--audio interfaces may under-expose alignment cues.
We hope EcoG-Bench will drive progress in both models and interfaces that explicitly represent and exploit fine-grained audio--visual timing for deictic collaboration.

\bibliographystyle{splncs04}
\bibliography{main}
\end{document}